%% file: main.tex
\documentclass{style/naturep}

\usepackage{authblk}
\usepackage{amssymb, amsmath, graphicx, algorithmicx, algorithm, algpseudocode}
\usepackage{bbm, lineno, geometry, ragged2e, setspace, longtable, hyperref}
\usepackage{cleveref, anyfontsize, float, booktabs, multirow, xcolor, tabularx, caption, subcaption, geometry}
\usepackage{enumitem, kantlipsum}
\usepackage{array}  
\usepackage{tabularx}

\title{\raggedright{\textbf{Accelerating Data Processing and Benchmarking of AI Models for Pathology}}}

\author[1,2,3,4]{Andrew Zhang}
\author[1,2,3]{Guillaume Jaume}
\author[1,2,3,4]{Anurag Vaidya}
\author[1,2,3,5]{Tong Ding}
\author[1,2,3,*]{Faisal Mahmood}

\affil[1]{Department of Pathology, Brigham and Women's Hospital, Harvard Medical School, Boston, MA}
\affil[2]{Department of Pathology, Massachusetts General Hospital, Harvard Medical School, Boston, MA}
\affil[3]{Cancer Program, Broad Institute of Harvard and MIT, Cambridge, MA}
\affil[4]{Health Sciences and Technology, Harvard-MIT, Cambridge, MA}
\affil[5]{Harvard John A. Paulson School of Engineering and Applied Sciences, Harvard University, Cambridge, MA}
\affil[*]{\textbf{Corresponding author}: Faisal Mahmood (FaisalMahmood@bwh.harvard.edu)}

\makeatletter
\let\saved@includegraphics\includegraphics
\AtBeginDocument{\let\includegraphics\saved@includegraphics}
\makeatother

\begin{document}

\begin{spacing}{1.2}
\maketitle

\input{sec/0_abstract}


\input{sec/1_intro}
\input{sec/2_trident}
\input{sec/3_pathobench}

\input{sec/4_conclusion}

\end{spacing}


\clearpage
\setcounter{figure}{0}
\renewcommand{\figurename}{Extended Data Figure}

\begin{nolinenumbers}

\end{nolinenumbers}

\begin{nolinenumbers}
\section*{References} 
\vspace{2mm}

\begin{spacing}{0.9}
\bibliographystyle{naturemag}
\bibliography{main}
\end{spacing}
\end{nolinenumbers}

\end{document}

%% file: sec/0_abstract.tex

\noindent Advances in foundation modeling have reshaped computational pathology. However, the increasing number of available models and lack of standardized benchmarks make it increasingly complex to assess their strengths, limitations, and potential for further development. To address these challenges, we introduce a new suite of software tools: \href{https://github.com/mahmoodlab/trident/}{\textbf{Trident}} for whole-slide image processing, \href{https://github.com/mahmoodlab/patho-bench}{\textbf{Patho-Bench}} for foundation model benchmarking, and \href{https://huggingface.co/datasets/MahmoodLab/Patho-Bench}{\textbf{Patho-Bench tasks}} with clinically relevant tasks sourced from public data. We anticipate that these resources will promote transparency, reproducibility, and continued progress in the field.

%% file: sec/1_intro.tex
\clearpage
\noindent\textbf{\large{Introduction}} \label{sec:intro}

The scale of available histology data is rapidly expanding as more medical institutions transition to fully digitized workflows, supported by large-scale retrospective tissue slide scanning. Many labs and institutions now have access to petabytes of data accounting for millions of diagnostic slides. This shift has significantly advanced AI applications in pathology, evolving from early studies with hundreds of slides\cite{bejnordi2017diagnostic} to datasets with tens of thousands\cite{bulten2022artificial}, and now, with the advent of foundation models (FMs)\cite{chen2024towards,vorontsov2024foundation,xu2024whole}, to training on millions. Foundation models offer a shared basis for developing task-specific models at minimal cost, enabling adaptation to various clinically relevant tasks such as predicting histologic subtypes, molecular biomarkers, and treatment response directly from the tissue morphology\cite{kather2020pan,lu2021ai}.

However, current open-source tools for whole-slide image (WSI) processing are not designed for scaling to very large repositories with support for multiple stains including hematoxylin and eosin (H\&E), immunohistochemistry and special stains\cite{pocock2022tiatoolbox,elnahhas2025stamp}. With a global acceleration in the number of publicly available foundation models in pathology, it also becomes increasingly complex to understand their strengths and weaknesses compared to existing models. This is compounded by the lack of reusable open-source code with standardized train-test data splits on diverse downstream tasks. 

To advance the field, the community needs new tools for foundation model evaluation based on very large benchmarks. Model assessment must be based on several metrics aggregated across many tasks, hyperparameter combinations and evaluation strategies, such as linear probing, supervised fine-tuning, and case retrieval. To address these challenges, we release a set of software packages and downstream tasks that aim to standardize foundation model benchmarking: \textbf{Trident}, a package for whole-slide image processing with support for state-of-the-art patch-level and slide-level foundation models\cite{wang2024chief,shaikovski2024prism,vaidya2024amolecular,ding2024titan}, \textbf{Patho-Bench}, a library for benchmarking FMs under several evaluation strategies, and \textbf{Patho-Bench data splits}, with labels for 42 clinically relevant pathology tasks.

%% file: sec/2_trident.tex
\noindent\textbf{\large{Accelerating WSI processing with Trident}} \label{sec4}

We introduce Trident\footnote{\url{https://github.com/mahmoodlab/trident/}}, a Python package for processing WSIs using pretrained foundation models for pathology. Building on the widely adopted CLAM toolbox\cite{luDataefficientWeaklySupervised2021a}, Trident addresses key limitations of its predecessor, including limited error handling, lack of support for recent foundation models at both patch and slide levels, and challenges in scaling to large repositories. To bridge this gap, Trident offers: (i) support for most WSI formats across multiple stains, (ii) a robust tissue-vs-background segmentation pipeline, (iii) access to 18 popular foundation models via a unified API, and (iv) scalable batch processing modules capable of handling thousands of WSIs.



\subsection{Tissue vs. background segmentation.}
Tissue segmentation removes background regions to minimize unnecessary downstream processing. Existing packages often use image processing techniques such as binary or Otsu thresholding of pixel intensity, which typically require manual tuning. Moreover, these methods struggle to generalize beyond H\&E staining and fail to effectively separate tissue from noise and artifacts, such as penmarks or bubbles. Instead, Trident uses a segmentation model based on DeepLabV3 pretrained on the COCO dataset\cite{jaumeHEST1kDatasetSpatial2024}. The tissue segmentation can also be edited in QuPath\cite{bankheadQuPathOpenSource2017} to correct mistakes or restrict slide processing to a region-of-interest.


\subsection{Tissue patching.}
Post-tissue segmentation, the patching step divides the tissue-containing regions into individual image patches for later processing by a patch encoder. For efficiency, only patch coordinates are extracted, with patch images being loaded on demand during the feature extraction step. The patching step is specified using two parameters: patch size and magnification (i.e., 256 $\times$ 256-pixel patches at 20$\times$ magnification).
As retrieving metadata about the image resolution and magnification may be complex, Trident uses several heuristics to determine the raw magnification level. If all methods fail, users can manually provide the pixel resolution.


\subsection{Feature extraction.}
Trident provides model factories for easily loading pretrained patch and slide encoders and unifying inference. Trident provides off-the-shelf support for 13 publicly released patch encoders, including UNI\cite{chen2024towards}, CONCH\cite{luVisuallanguageFoundationModel2024}, Virchow\cite{vorontsov2024foundation}, among others (\Cref{tab:foundation_models}). We also support five slide encoder FMs, including Threads\cite{vaidya2024amolecular}, Titan\cite{ding2024titan}, PRISM\cite{shaikovski2024prism}, Prov-GigaPath\cite{xu2024whole} and CHIEF\cite{wang2024chief} (\Cref{tab:foundation_models}). As new models are released, they can easily be integrated into Trident with minimal effort. These pretrained FMs can either be used to extract patch and slide-level features using the provided scripts or imported into custom pipelines for inference or finetuning.

\begin{table}[t]
    \centering
    \caption{\textbf{Publicly available patch-level and slide-level foundation models supported by Trident.}
    }
    \begin{tabular}{cc}
    \toprule
    \textbf{Patch encoders} & \textbf{Slide encoders} \\
    \midrule
    UNI \cite{chen2024towards}                   & Threads \cite{vaidya2024amolecular}                \\
    UNIv2 \cite{chen2024towards}                   & Titan \cite{ding2024titan}                 \\
    CONCH \cite{luVisuallanguageFoundationModel2024}                  &  PRISM \cite{shaikovski2024prism}                \\
    CONCHv1.5 \cite{luVisuallanguageFoundationModel2024}              &  CHIEF \cite{wang2024chief}                 \\
    Virchow \cite{vorontsov2024foundation}                &  Prov-Gigapath \cite{xu2024whole}                 \\
    Virchow2 \cite{zimmermannVirchow2ScalingSelfSupervised2024}              &  Mean pooling         \\
    Phikon \cite{filiotScalingSelfSupervisedLearning2023}                 &                        \\
    Phikon-v2 \cite{filiotPhikonv2LargePublic2024}              &                        \\
    Prov-GigaPath \cite{xu2024whole}          &                        \\
    H-Optimus-0 \cite{saillardHOptimus02024}           &                        \\
    MUSK \cite{xiangVisionLanguageFoundation2025}                   &                        \\
    CTransPath \cite{wangTransformerbasedUnsupervisedContrastive2022}             &                        \\
    ResNet50-ImageNet \cite{heDeepResidualLearning2015}      &                        \\
    \bottomrule
    \end{tabular}
    \label{tab:foundation_models}
\end{table}


%% file: sec/3_pathobench.tex
\clearpage
\noindent\textbf{\large{Standardizing benchmarking with Patho-Bench}} \label{sec5}

We introduce Patho-Bench\footnote{\url{https://github.com/mahmoodlab/patho-bench}}, a Python package for large-scale model evaluation, which can manage thousands of experiments with efficient parallelism. In addition, we publicly release a unified set of tasks with clean labels and predefined train-test splits\footnote{\url{https://huggingface.co/datasets/MahmoodLab/patho-bench}}. Patho-Bench is the most extensive and diverse public benchmark for computational pathology released to date.

\subsection{Downstream tasks and data splits.}
We curated canonical train-test splits for 42 publicly available WSI-level and patient-level tasks, which we categorized into six families: morphological subtyping, tumor grading, molecular subtyping, mutation prediction, treatment response and assessment, and survival prediction. A description of each task family is provided in \Cref{tab:tasks}. Detailed information on each dataset and task is provided in \cite{vaidya2024amolecular}.

\begin{table}[ht]
    \centering
    \caption{\textbf{Overview of families of tasks in Patho-Bench (42 public tasks in total).}}
    \renewcommand{\arraystretch}{1.2}
    \begin{tabularx}{\textwidth}{lXc}
        \toprule
        \textbf{Task Family} & \textbf{Description} & \textbf{\# Tasks} \\
        \midrule
        Morphological subtyping & Classifying different disease subtypes & 4 \\
        Tumor grading & Assigning a grade based on cellular differentiation and growth patterns & 2 \\
        Molecular subtyping & Predicting molecular alterations as tested with immunohistochemistry & 3 \\
        Mutation prediction & Predicting genetic mutations in tumors as tested with next-generation sequencing & 21 \\
        Treatment response and assessment & Evaluating how patients respond to treatment & 6 \\
        Survival prediction & Predicting time-to-event for patient survival outcomes & 6 \\
        \bottomrule
    \end{tabularx}
    \label{tab:tasks}
\end{table}

Each task is associated with two artifacts: A CSV file and a YAML file. The CSV file contains one row per slide, with columns indicating the patient and slide identifier, task label, and train-test assignments for each fold. The YAML file contains additional task metadata such as whether it is a patient-level or slide-level task, the number of samples, number of folds, and task-dependent canonical performance metric (e.g., balanced accuracy, area under the receiver operating characteristic curve, quadratic weighted kappa, or concordance index). Most tasks use 5-fold cross-validation, while some tasks use 50-fold Monte Carlo sampling due to having very few samples. For a small number of tasks that already have canonical single-fold train-test splits reported in the literature, we use their official splits. Otherwise, we maintain a train-test ratio of 80\%:20\% in each split. We have deliberately refrained from assigning validation samples in the public splits, leaving it up to the user to decide whether a validation set is suitable for their task and how to reassign any training samples toward a validation set. 

\subsection{Evaluation frameworks.}
In Patho-Bench, models can be evaluated using three different parametric evaluation strategies (linear probing, Cox proportional-hazards regression, and supervised finetuning) and one non-parametric one (case retrieval). Supervised finetuning uses frozen patch-level features, while the remaining evaluation frameworks use frozen slide- or patient-level features. All evaluation frameworks in Patho-Bench use Trident-extracted patch features. Patho-Bench incorporates the slide-level feature extraction abilities of Trident while also adding the ability to extract patient-level features using the data split CSVs.

\subsection{Parallelization.}
A major challenge of running large benchmarks is the combinatorial experimentation space. For instance, running benchmarks for five FMs across 50 tasks and three evaluation frameworks per task adds up to already 750 individual experiments. This does not even consider hyperparameter sweeps, multiple folds of training and testing, or few shot variations of each task, where benchmarking a single model can easily scale to hundreds of thousands of training and testing runs. It is infeasible to loop over all possible experimental configurations in a serial manner, and manually parallelizing experiments takes a significant amount of effort and is prone to human error. To mitigate this challenge, Patho-Bench can run a set of experiments using task-level parallelization. The user can easily define a parallelization strategy in a configuration file, which will automatically launch and monitor all requested experiments in the terminal using the popular Linux utility Tmux. For experiments requiring GPUs, Patho-Bench will automatically perform load balancing across available GPUs. At the end of a sweep, all experiment results are automatically gathered into a single output file.

In designing Patho-Bench, we acknowledged that users are likely to prefer varying levels of complexity and that some users may not need the full parallelized implementation. Therefore, for each type of evaluation, we expose both high-level scripts for high-throughput experiment sweeps as well as low-level modules that enable users to run single experiments. The implementation is highly modular, making it easy to reuse and adapt for specialty use cases. Users can also incorporate Patho-Bench task labels and data splits into their own evaluation pipelines.

%% file: sec/4_conclusion.tex
\noindent\textbf{\Large{Conclusion}} 

Trident and Patho-Bench are a significant step toward better transparency and reproducibility in computational pathology.
Compared with existing offerings, our codebases are optimized for FM development and conform to the principle that code simplicity and reusability are more important than handling the entire experiment lifecycle end-to-end. Moving forward, our hope is that both new and existing labs working on pathology foundation models will find these resources valuable as a common starting ground. We believe that collaboration is important for accelerating scientific progress, and therefore welcome contributions to these open-source repositories from the research community.
